\documentclass[10pt,twocolumn,letterpaper]{article}

\usepackage{titling}
\usepackage{iccv}
\usepackage{times}
\usepackage{epsfig}
\usepackage{graphicx}
\usepackage{amsmath}
\usepackage{amssymb}

\usepackage{algorithm}
\usepackage{listings}
\usepackage{bm}
\usepackage{booktabs}
\usepackage{multirow}
\usepackage{subcaption}
\usepackage{stfloats}

\newcommand{\tabitem}{~~\llap{\textbullet}~~}

\usepackage[pagebackref=true,breaklinks=true,letterpaper=true,colorlinks,bookmarks=false]{hyperref}

\iccvfinalcopy 

\ificcvfinal\fi

\begin{document}
\pdfsuppresswarningpagegroup=1

\title{Pi-NAS: Improving Neural Architecture Search by Reducing Supernet Training Consistency Shift}

\author{
Jiefeng Peng$^{1,2}$\thanks{Jiefeng Peng and Jiqi Zhang are co-first authors and share equal contributions. Their names are listed in alphabetical order.}, \quad Jiqi Zhang$^{1}$\footnotemark[1], \quad Changlin Li$^{3}$, \quad Guangrun Wang$^{4}$\thanks{Corresponding Author.},\\ \quad Xiaodan Liang$^{1}$, \quad Liang Lin$^{1}$\\
\small$^1$Sun Yat-sen University\quad \small$^2$DarkMatter AI Research \\ \quad \small$^3$GORSE Lab, Dept. of DSAI,
Monash University \quad \small$^4$University of Oxford\\
{\tt\small \{jiefengpeng,wanggrun,xdliang328\}@gmail.com,}\\{\tt\small zhangjq49@mail2.sysu.edu.cn, changlin.li@monash.edu, linliang@ieee.org}
\vspace{-12pt}}

\maketitle
\ificcvfinal\thispagestyle{empty}\fi

\begin{abstract}\vspace{-5pt}
Recently proposed neural architecture search (NAS) methods co-train billions of architectures in a supernet and estimate their potential accuracy using the network weights detached from the supernet. However, the ranking correlation between the architectures' predicted accuracy and their actual capability is incorrect, which causes the existing NAS methods' dilemma. We attribute this ranking correlation problem to the supernet training consistency shift, including \textbf{feature shift} and \textbf{parameter shift}. Feature shift is identified as dynamic input distributions of a hidden layer due to random path sampling. The input distribution dynamic affects the loss descent and finally affects architecture ranking. Parameter shift is identified as contradictory parameter updates for a shared layer lay in different paths in different training steps. The rapidly-changing parameter could not preserve architecture ranking. We address these two shifts simultaneously using a nontrivial supernet-$\Pi$ model, called $\Pi$-NAS. Specifically, we employ a supernet-$\Pi$ model that contains cross-path learning to reduce the feature consistency shift between different paths. Meanwhile, we adopt a novel nontrivial mean teacher containing negative samples to overcome parameter shift and model collision. Furthermore, our $\Pi$-NAS runs in an unsupervised manner, which can search for more transferable architectures. Extensive experiments on ImageNet and a wide range of downstream tasks (e.g., COCO 2017, ADE20K, and Cityscapes) demonstrate the effectiveness and universality of our $\Pi$-NAS compared to supervised NAS. See Codes\footnote{Code: \url{https://github.com/Ernie1/Pi-NAS}}.
\end{abstract}

\vspace{-14pt}
\section{Introduction}\label{sec: intro}
\vspace{-5pt}

\begin{figure}[t]
\centering
\begin{subfigure}{0.47\textwidth}
  \centering
  \includegraphics[width=.9\linewidth]{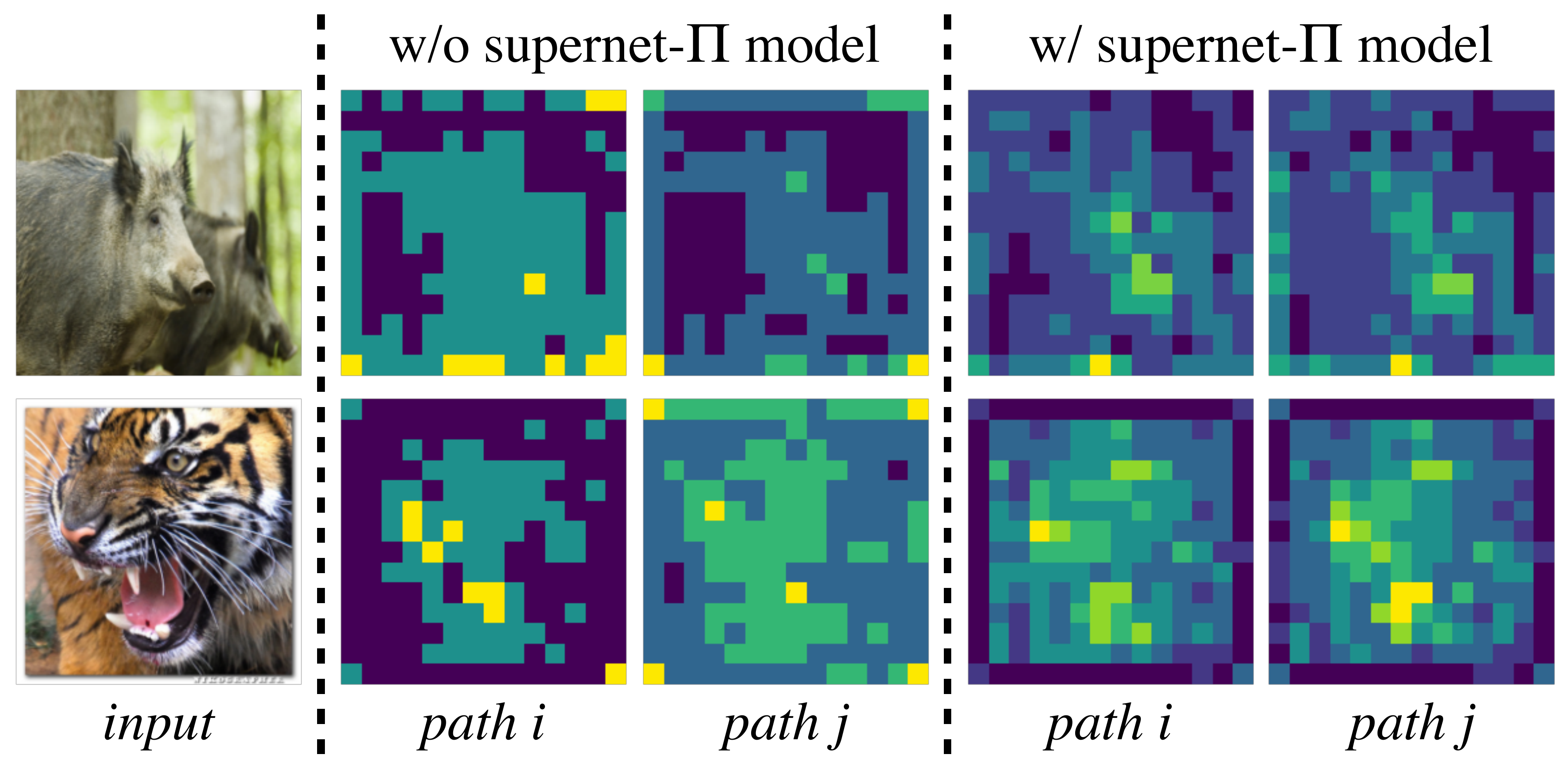}
  \vspace{-5pt}
  \caption{Feature shift. \textbf{Left:} Without the supernet-$\Pi$ model, there is a feature shift between different paths' feature maps. \textbf{Right:} With supernet-$\Pi$ model, the feature shift is significantly reduced.}
  \label{fig:feature_shift}
\end{subfigure}
\begin{subfigure}{0.47\textwidth}
   \centering
   \includegraphics[width=.9\linewidth]{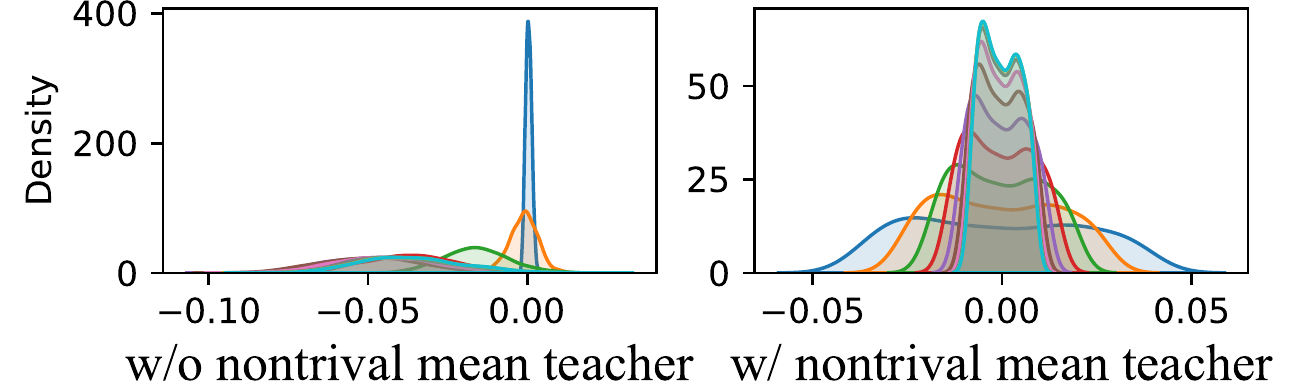}
   \vspace{-3pt}
   \caption{Parameter shift. Different colors represent the distribution of parameters in different iterations. \textbf{Left:} without our nontrivial mean teacher, the parameter has significantly varying distributions in training. \textbf{Right:} with our nontrivial mean teacher, the parameter shift is significantly reduced.}
   \label{fig:parameter_shift}
\end{subfigure}
\vspace{-11pt}
\caption{\small{Illustration of supernet training consistency shift.}}
\label{fig:intro}
\vspace{-15pt}
\end{figure}

Automatic neural architecture search (NAS) has been an intense longing in machine learning in the past four years. Early works use reinforcement learning \cite{zoph2016neural} or evolutionary algorithms \cite{real2017large} to discover high-performance architectures in the search space. The searching procedure usually costs thousands of GPU days for large datasets, as each sampled architecture needs training from scratch. Recently, to alleviate the heavy burden, weight sharing NAS methods \cite{Chu2019FairNA_arxiv,cai2018proxylessnas,guo2020single,wu2019fbnet,li2020improving,li2020block,chu2019scarletnas, Liu2018DARTSDA, dong2019searching, akimoto2019adaptive, brock2017smash} are widely used, where candidate architectures share weights and train simultaneously in a supernet\footnote{A supernet is an over-parameterized network that integrates the entire search space. Each architecture within the search space corresponds to a supernet's sub-net capturing the required operations.}. After training, a candidate subnet's weights detached from the supernet are used to predict its actual performance. Despite the remarkable progress in efficiency, weight-sharing NAS's effectiveness is still unstable, i.e., it has a low ranking correlation between candidates' actual accuracies and accuracies estimated in supernet. In short, inaccurate architecture ranking is an inevitably critical problem in today's NAS.

In this paper, we attribute the ranking correlation problem to the supernet training consistency shift, including \emph{feature shift} and \emph{parameter shift}. \textbf{Feature shift} is identified as dynamic input distributions of a hidden layer. Specifically, a given layer's input feature maps always have an uncertain distribution due to random path sampling (see Figure \ref{fig:feature_shift}, left). This distribution uncertainty can hurt the architecture ranking correlation. Precisely, we can use the loss to measure the architecture accuracy, and we can link the accuracy ascent to gradient descent. Based on the back-propagation rule, a stable input distribution can guarantee a good ranking correlation. In contrast, the input distribution dynamic affects the loss descent and finally affects architecture ranking. \textbf{Parameter shift} is identified as contradictory parameter updates for a given layer. In supernet training, a given layer will always be present in different paths from iteration to iteration (see Figure \ref{fig:parameter_shift}, left).  The parameter in this layer may have a contradictory update from iteration to iteration. These unstable updates lead to varying parameters' distributions, hurting the architecture ranking correlation in two ways. On the one hand, stable parameters can ensure a correct loss descent and guarantee an accurate architecture ranking, while frequent parameter change could not preserve architecture ranking. On the other hand, varying parameters can also result in a feature shift, further hurting architecture ranking correlation. In summary, both feature shift and parameter shift can hurt the architecture ranking correlation. \textbf{Detailed experimental analysis} in Section \ref{sec:exp} provide solid evidence to support this analysis.

Motivated by consistency regularization methods \cite{laine2016temporal, tarvainen2017mean}, we propose a nontrivial supernet-$\Pi$ model, called $\Pi$-\textbf{NAS}, to reduce these two shifts simultaneously. Specifically, to cope with the \emph{feature shift}, we propose a novel supernet-$\Pi$ model. We evaluate each data point through two randomly sampled paths, then apply a consistency cost between the two predictions to penalize the feature consistency shift between different paths. As shown in Figure \ref{fig:feature_shift} (right), our method can significantly reduce the feature shit and thus can improve the architecture ranking correlation. To address the \emph{parameter shift}, we propose a novel nontrivial mean teacher model by maintaining an exponential moving average of weights in supernet teacher. Although a mean teacher can stabilize the parameters in single network training, it could be trapped in a trivial solution and lead to a model collision in supernet training. Our nontrivial mean teacher novelly contains appropriate negative samples to avoid such a model collision. An impressive result of our method in reducing the parameter shift is shown in Figure \ref{fig:parameter_shift} (right). In brief, our $\Pi$-NAS can reduce the supernet training consistency shift and thus improve the architecture ranking, which is critical for NAS's effectiveness.

One by-product that could not be ignored is that our $\Pi$-NAS runs in an unsupervised manner, which has an additional gain that existing supervised NAS methods do not have. Concretely, similar to unsupervised representation learning that can learn general features, our $\Pi$-NAS can search for more transferrable and universal architectures than supervised NAS counterparts.

Since the ``good architectures'' in previous NAS search spaces usually have considerable computation complexity, using these search spaces for evaluation lacks interpretability. To evaluate our $\Pi$-NAS, we design a nontrivial search space based on 16-layer ResNet-50. Our searched models on this space achieve a state-of-the-art top-1 accuracy of 81.6\% on ImageNet, surpassing ResNeSt-50 by 0.5\% with comparable computation cost. We also validate $\Pi$-NAS on NAS-Bench-201 with CIFAR-10, beating state-of-the-art NAS methods and verifying our method’s effectiveness. In addition, our $\Pi$-NAS models keep state-of-the-art on many downstream tasks (e.g., COCO 2017 detection and segmentation, ADE20K segmentation, and Cityscapes segmentation), demonstrating the universality of our $\Pi$-NAS.

Overall, this paper makes three contributions.

\begin{itemize}
\vspace{-6pt}
\item We attribute the inaccurate architecture ranking to the supernet training consistency shift, including feature and parameter shifts. Then we provide a detailed empirical analysis of how these two shifts are making NAS methods ineffective.

\vspace{-6pt}
\item We propose a $\Pi$-NAS method with two key components, i.e., a supernet-$\Pi$ model and a nontrivial mean teacher, to address feature shift and parameter shift, respectively. Notably, our nontrivial mean teacher model introduces appropriate negative samples to avoid being trapped in a trivial solution. 

\vspace{-6pt}
\item Our $\Pi$-NAS method shares the merit of unsupervised representation learning, i.e., the universality property. We can search for architectures that are more transferrable and universal than supervised NAS methods. Substantial empirical results are obtained on ImageNet and a wide range of downstream tasks to demonstrate the effectiveness and universality of our $\Pi$-NAS.
\vspace{-3pt}
\end{itemize}

\section{Related Work}\label{sec:related}
\vspace{-3pt}

\noindent\textbf{Neural Architecture Search (NAS).} NAS has attracted increasing research attention in recent years. Early NAS works~\cite{zhong2018practical, chen2018reinforced, negrinho2017deeparchitect,zoph2016neural,baker2016designing,real2017large,tan2019mnasnet} consume a huge amount of computation resources to train thousands of candidate models from scratch while using an agent (an RNN controller or evolution algorithm) to explore better-performing architectures in the search space. To alleviate the computational overhead caused by the training process, researchers starts to share the weights among candidate architectures~\cite{Chu2019FairNA_arxiv,cai2018proxylessnas,guo2020single,wu2019fbnet,li2020improving,li2020block,chu2019scarletnas, Liu2018DARTSDA, dong2019searching, akimoto2019adaptive, brock2017smash}. Gradient-based weight sharing methods~\cite{Liu2018DARTSDA,cai2018proxylessnas, wu2019fbnet,ZhangLPCGS20} jointly optimize the shared network parameters and the architecture choosing factors by gradient descent. In one-shot methods~\cite{guo2020single,Chu2019FairNA_arxiv, brock2017smash, bender2018understanding, li2020block}, the supernet is first optimized with path sampling, and then sub-models are sampled and evaluated with the weights inherited from the supernet. Despite the acceleration of weight sharing, these approaches still suffer a critical issue on their effectiveness~\cite{bender2018understanding, Chu2019FairNA_arxiv, li2020improving}. Existing attempting on solving this issue includes ensuring optimization fairness among all child models ~\cite{Chu2019FairNA_arxiv}, reducing the search space greedily during training~\cite{li2020improving}, modularizing the large search space into blocks using an intermediate knowledge distillation~\cite{li2020block} and constraining the subnet optimization to prevent multi-model forgetting~\cite{zhang2020overcoming,zhang2020one}. Recently, unsupervised NAS methods are also starting to attract research interest ~\cite{liu2020labels,yan2020does,li2021bossnas,zhang2021randomlabelnas,wang2021Joint_tnnls}.

\noindent\textbf{Reducing Consistency Shift.} \textit{Feature shift} is represented as the instability of the network to the perturbation of an input image. Penalizing the consistency shift can help develop the network's tolerance to incorrect labels and improve the classification accuracy in semi-supervised learning~\cite{bachman2014learning,sajjadi2016regularization,laine2016temporal,tarvainen2017mean,zhai2019s4l,xie2019unsupervised,oliver2018realistic,berthelot2019mixmatch,xie2020self,wang2017deep_iccv,wang2020weakly_tnnls,wang2020Smoothing_cvpr}. \cite{laine2016temporal} proposes $\Pi$-model to encourage consistent output for input with different augmentation and dropout, and extend the $\Pi$-model by temporal ensembling the network's output for each input, to retain the consistency of the outputs. \textit{Parameter shift} is represented as the instability of network parameters. To address the parameter shift, a mean teacher model ~\cite{tarvainen2017mean} refines the temporal ensembling by averaging the model weights rather than outputs, which has also been used to stabilize weight sharing training \cite{LiWWLLC21}. In this paper, we attribute NAS's inefficiency to incorrect architecture ranking caused by supernet training consistency shift, i.e., feature shift and parameter shift. Since $\Pi$ model is a classical tool to reduce feature shift, we propose a supernet-$\Pi$ model to address the feature shift. Our supernet-$\Pi$ model is a novel one as we use a novel formulation of cross-path learning. On the other hand, mean-teacher is widely adopted to reduce parameter shift because it can reliably reduce the implausible uncertainties. Hence, we introduce mean teacher to address our parameter shift. Although a mean teacher can be employed to stabilize the parameters in single network training, it could be trapped in a trivial solution and lead to a model collision in supernet training. Our nontrivial mean teacher novelly contains appropriate negative samples to avoid such a model collision. In summary, our method is a nontrivial NAS method aiming at closing the supernet training consistency shift, but not a straightforward combination of the NAS and $\Pi$ model and mean teacher.

\noindent\textbf{Contrastive Learning.} Recent contrastive learning-based methods have brought a leap in unsupervised representation learning~\cite{oord2018representation,wu2018unsupervised,hjelm2018learning,tian2019contrastive,zhuang2019local,he2020momentum,chen2020simple,wang2021Solving_iccv}. Being cast as either the dictionary look-up task~\cite{wu2018unsupervised,he2020momentum} or the consistent learning task~\cite{tian2019contrastive,chen2020simple}, these methods learn discriminative representations by bringing the representation of different views of the same image closer and spreading representations of views from different images apart. MoCo~\cite{he2020momentum, chen2020improved} uses an exponential moving average (EMA) encoder to generate predictions and keep a large bank of the historical predictions as the negative samples. In BYOL~\cite{grill2020bootstrap}, the online network with a predictor is trained to be consistent with the EMA target network without requiring negative pairs. However, straightforward applying the technique from contrastive learning to NAS could be either unnecessary or unsuccessful. Due to the training consistency shift, there will be a feature shift in a pair of samples in contrastive learning, especially in negative sample pairs. This makes the supernet optimization unstable and hard to convergent. In contrast, our $\Pi$-NAS contains a cross-path training formulation that can satisfactorily address the feature shift problem.

\begin{figure*}[t]
\begin{center}
\vspace{-33pt}
\includegraphics[width=1\linewidth]{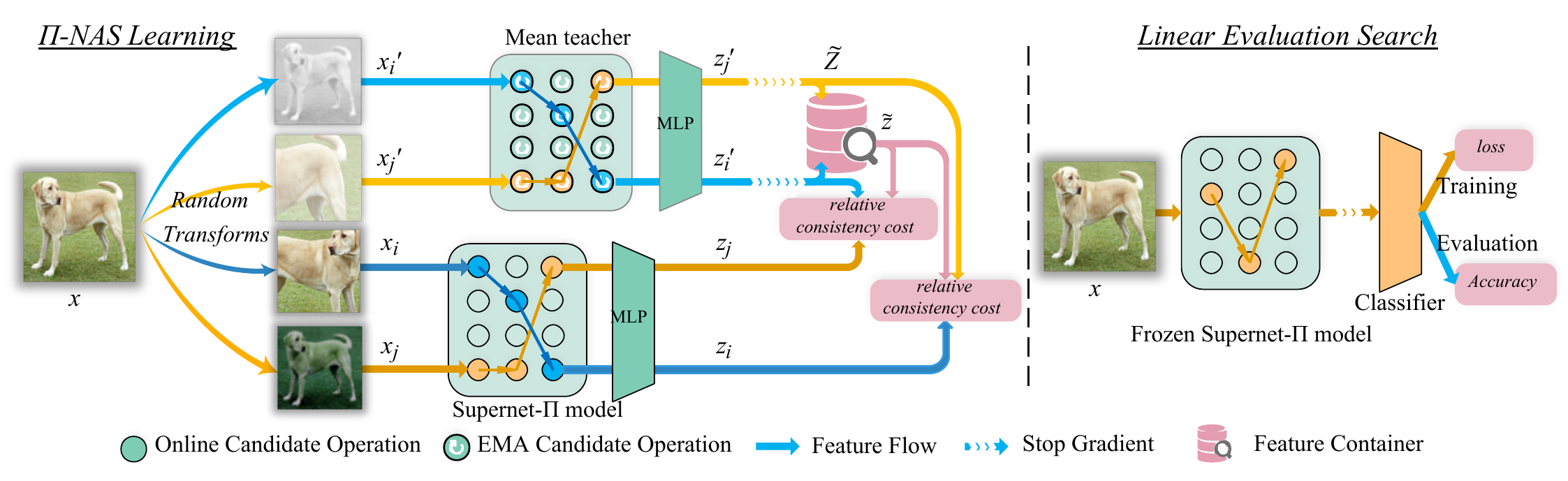}
\end{center}
\vspace{-18pt}
  \caption{An overview of proposed $\Pi$-NAS method. In \textit{$\Pi$-NAS learning}, the input image is transformed into four different views that are then separately routed through the supernet-$\Pi$ model and nontrivial mean teacher to calculate relative consistency cost with negative targets. The outputs of the mean teacher are saved in the container to serve as negative targets in the future. In \textit{Linear Evaluation Search}, the classifier is trained analogous to linear evaluation, and the accuracy is used as the metrics of architecture search.}\label{fig:consistency regularization}
  \vspace{-11pt}
\end{figure*}

\vspace{-3pt}
\section{Methodology}\label{sec:method}
\vspace{-3pt}
We first briefly introduce the dilemma of NAS, i.e., inaccurate architecture ranking, then attribute incorrect architecture ranking to the supernet training consistency shift, including feature shift and parameter shift. Then, we propose a nontrivial supernet-$\Pi$ model with two key components, i.e., a supernet-$\Pi$ model and a nontrivial mean teacher, to address feature shift and parameter shift, respectively. At last, we search promising architecture in linear evaluation.

\vspace{-3pt}
\subsection{Dilemma of NAS}\label{sec:dilemma}
\vspace{-3pt}
\noindent\textbf{Inaccurate architecture ranking.} Let $\mathcal{A}$ denote the architecture search space. $\alpha \in \mathcal{A}$ and $\omega_\alpha$ are the network architecture and the network weights, respectively. As mentioned above, NAS aims to find an optimal pair ($\alpha^*$, $\omega_\alpha^*$) such that the model performance is maximized in search space $\mathcal{A}$.
The searching procedure can be formulated as two subproblems. The first one is a \emph{architecture training} that trains the network weights of given architectures. The second one is an \emph{architecture search} that searches for an architecture with the best performance if trained. As training each architecture from scratch to convergence is prohibitive in practice due to the high computation cost, recently, weight-sharing NAS was proposed. \cite{cai2018proxylessnas, guo2020single, wu2019fbnet, li2020improving, li2020block} propose to train for different candidates concurrently via a weight sharing strategy, encoding the search space $\mathcal{A}$ in an over-parameterized supernet. Thus, all candidate architectures can inherit their weights immediately from the supernet. However, the proxy weights borrowed from the supernet do not adequately indicate network weights trained from scratch to convergence, as each subgraph is not fairly and sufficiently optimized in supernet. This may lead to a low ranking correlation between the candidates' predicted accuracy and their actual capability, which causes the ineffectiveness of architecture search. We identify this as the dilemma of NAS.

\textbf{Supernet training consistency shift.} So, what causes the dilemma of NAS? In this paper, we attribute the inaccurate architecture ranking to the supernet training consistency shift, which contains feature shift and parameter shift.

\textit{Feature shift} is identified as dynamic input distributions of a hidden layer. Let $\mathbf{x}_l$ denote the input of layer $l$ and $\mathbf{y}_l$ denote its output. $\mathbf{w}_l$ is its network weights. Since the final architecture accuracy is inaccessible during the training, we use the loss  $\mathcal{L}$ to measure the architecture accuracy, and the accuracy ascent can be connected to the loss descent. According the chain rule of differentiation in the back-propagation algorithm, we have: $\frac{\partial\mathcal{L}}{\partial\mathbf{w}_l} = \frac{\partial\mathcal{L}}{\partial\mathbf{y}_l} \frac{\partial\mathbf{y}_l}{\partial\mathbf{w}_l} = \frac{\partial\mathcal{L}}{\partial\mathbf{y}_l} \mathbf{x}_l$. This indicates architecture ranking-preserving is highly dependent on the inputs $\mathbf{x}_l$. But for a given layer $l$, due to random path sampling in supernet, the preceding path varies, and the input $\mathbf{x}_l$ also varies. We thus should guarantee a stable $\mathbf{x}_l$ to preserve a good architecture ranking correlation. Otherwise, an input distribution dynamic impacts the loss descent and finally affects architecture ranking.

\textit{Parameter shift} is identified as contradictory parameter updates for a given layer. In supernet training, a given layer $l$ will always be present in different paths from iteration to iteration.  Its weights may have a contradictory update from iteration to iteration, i.e., $\mathbf{w}_l^{t+1} \leftarrow\mathbf{w}_l^t - \frac{\partial{\mathcal{L}^t}}{\partial\mathbf{w}_l^t}$. The rapidly-varying $\mathbf{w}_l$ will hurt the architecture ranking correlation in two ways. On the one hand, the loss descent is not only connected to $\frac{\partial{\mathcal{L}}}{\partial\mathbf{w}_l}$ but is also connect $\mathbf{w}_l - \frac{\partial{\mathcal{L}}}{\partial\mathbf{w}_l}$. This indicates that stable parameters can ensure a correct loss descent and guarantee an accurate architecture ranking, while frequently-varying parameters could not preserve architecture ranking. On the other hand, since the input $\mathbf{x}_l$ is generated by the network weights of the previous layers, varying parameters can also result in a feature shift, which further hurts architecture ranking correlation. 

In summary, both feature shift and parameter shift can hurt the architecture ranking correlation, further making NAS methods ineffective. \textbf{Detailed experimental analysis} in Section \ref{sec:exp} provide evidence to support this analysis.

\vspace{-2pt}
\subsection{{$\Pi$-NAS: A Nontrivial Supernet-${\Pi}$ Model}}\label{sect:ccl}
\vspace{-3pt}
As discussed, reducing the supernet training consistency shift can alleviate the dilemma of NAS. In the following, we design a novel and effective nontrivial supernet-$\Pi$ model, including a supernet-$\Pi$ model and a nontrivial mean teacher model, to address feature shift and parameter shift, respectively. Our $\Pi$-NAS can successfully preserve the architecture ranking and thus improve NAS's effectiveness.

\noindent{\textbf{Supernet-${\Pi}$ model.}}
To guarantee a stable input distribution, we are devoted to penalizing the inconsistency between the same input predictions through different sampled paths. Motivated by a $\Pi$ model, we evaluate data point $x$ through two randomly sampled paths, denoted as path $i$ and $j$, to get its representations $\{z_i, z_j, z'_i, z'_j\}$. Note that we obtain representations $z$ and $z'$ with different views of 
augmentation, i.e., $z=f(x)$ and $z'=f'(x)$, where $f$ and $f'$ are mapping functions of the supernet model. Without loss of generality, we define $f$ and $f'$ as the student/teacher models. Normally, the student and the teacher are identical.

After obtaining evaluations of the same input $x$, we define a cross-path consistency cost as follow:\begin{small}\begin{equation}\label{eq: generic consistency cost}
\mathcal{L}_{Con} = -\mathop{\mathbb{E}}\limits_{X}[\mathcal{D}(z_i, z'_j) + \mathcal{D}(z_j, z'_i)]
\end{equation}\end{small}where $X$ and $\mathcal{D}$ denote a training data set and a consistency metric, respectively. Figure \ref{fig:consistency regularization} shows a pipeline of our supernet-$\Pi$ model with cross-path learning. By minimizing Eqn. \ref{eq: generic consistency cost}, one could reduce the feature consistency shift caused by different random paths and thus stabilize the distributions of input features of a hidden layer.

In brief, we formulate our method under the $\Pi$ framework with cross-path learning, i.e., supernet-$\Pi$ model. Extensive experiments show a remarkable improvement in the architecture ranking correlation.

\noindent{\textbf{Nontrivial mean teacher model.}}
Besides addressing feature shift, we also intend to reduce parameter shift by smoothing parameter updates from iteration to iteration. Inspired by mean-teacher \cite{tarvainen2017mean}, we propose to maintain an exponential moving average weights for teacher model rather than barely replicate from student model in supernet-$\Pi$ model training. Formally, we denote $\mathcal{W}_t$ as parameters of student mapping function $f$ at training step $t$. Then, weights of mean teacher model $f'$ can be defined as:\begin{small}\begin{equation}\label{eq: mean teacher}
\mathcal{W}'_t = \lambda\mathcal{W}'_{t-1} + (1-\lambda)\mathcal{W}_t
\end{equation}\end{small}where $\lambda \in [0,1]$ is a smoothing coefficient hyper-parameter.

Although the capability of a mean teacher to stabilize the parameters is obvious, it could be trapped in a trivial solution in the supernet-$\Pi$ model. Specifically, barely optimizing consistency loss might lead to model collapse. For example, representations that are constant across arbitrary inputs are always entirely consistent. To circumvent this problem, we introduce appropriate negative samples to our model, i.e., nontrivial mean teacher model. Formally, an additive consistency cost is:\begin{small}\begin{equation}\label{eq: additive consistency cost}
\mathcal{L}_{Add} = \mathop{\mathbb{E}}\limits_{X}\left[\mathop{\mathbb{E}}\limits_{\widetilde Z}[\mathcal{D}(z_i, \widetilde z) + \mathcal{D}(z_j, \widetilde z)]\right]
\end{equation}\end{small}where $\widetilde Z$ represents a whole collection of negative samples $\widetilde z$, and $\widetilde z \in \widetilde Z$. Note that negative samples $\widetilde z$ can be collected from our nontrivial mean teacher model by reusing the previous predictions (see the \emph{Feature Container} in Figure \ref{fig:consistency regularization}). A relative consistency cost can be written as: $\mathcal{L}_{Ref} = \mathcal{L}_{Con} + \mathcal{L}_{Add}$.

Since our target is to maximize the consistency metric between positive samples while minimizing the negative ones, we can formulate the optimization as the categorical cross-entropy of classifying the positive samples, with $\frac{\exp(\mathcal{D}(z, z'))}{\sum_{\widetilde Z}\exp({\mathcal{D}(z, \widetilde z)})+\exp(\mathcal{D}(z, z')}$ being the prediction. We model consistency metric $\mathcal{D}$ with dot-product similarity as $\mathcal{D}(z, z') =  z^T z'$. Thus the final loss function of $\Pi$-NAS is formulated as:\begin{small}\begin{equation}\label{eq: cross-path loss}
\vspace{-11pt}
\mathcal{L} = -\mathop{\mathbb{E}}\limits_{X}\big[
\log\frac{e^{z_i^T z'_j}}{\sum\limits_{\widetilde Z}e^{z_i^T \widetilde z} + e^{z_i^T z'_j}} \\ + \log \frac{e^{z_j^T z'_i}}{\sum\limits_{\widetilde Z}e^{z_j^T \widetilde z} + e^{z_j^T z'_i}}
\big].
\end{equation}\end{small}

\subsection{Linear Evaluation Search}
After optimizing the nontrivial supernet-$\Pi$ model with $\mathcal{W}$, an architecture search is conducted by evaluating the representation capability of candidates $\alpha$. Inspired by the standard linear evaluation protocol \cite{kolesnikov2019revisiting, hadsell2006dimensionality} using in self-supervised learning, we train a linear classifier on the top of the frozen representation, i.e., without updating the supernet parameters $\mathcal{W}$ nor the batch statistics. Specifically, the linear classifier $Fc$ is also optimized via a common weight sharing strategy. Then, we estimate the capability of the sub-model by its accuracy $\mathcal{R}_{val}$ on the validation set and search for the best performance:\begin{small}\begin{equation}\label{eq: linear evaluation search}\vspace{-11pt}
\alpha^* = \mathop{\arg\max}_{\alpha \in \mathcal{A}}{\mathcal{R}_{val}(Fc(\mathcal{W}_\alpha, \alpha; X, Y))}
\end{equation}\end{small}where $\mathcal{W}_\alpha$ is the sub-architecture $\alpha$'s parameters inherited directly from parameters $\mathcal{W}$. 

Thanks to $\Pi$-NAS learning and linear evaluation searching, our $\Pi$-NAS not only improves the search effectiveness but also shows the superiority in searching for more transferable and universal architectures. Finally, an overview of our $\Pi$-NAS is presented in Figure \ref{fig:consistency regularization}.
%
%

\section{Experiments}\label{sec:exp}

\subsection{Implementation Details}\label{sec:implementation_details}

\noindent\textbf{Search space and dataset.} We construct our supernet based on 16-layer ResNet-50 by replacing the residual bottleneck in each layer with 4 candidate \textit{Split-Attention} blocks~\cite{Zhang2020ResNeStSN} of radix $s$, cardinality $x$ and width $d$. Thus our search space $\mathcal{A}$ includes $4^{16}$ architectures.
\vspace{-2pt}
\begin{table}[h]
\centering
\begin{tabular}{llll}
\tabitem \textbf{\textit{Block0}}: $1s1x64d$ &&& \tabitem \textbf{\textit{Block1}}: $2s1x64d$     \\
\tabitem \textbf{\textit{Block2}}: $1s2x42d$ &&& \tabitem \textbf{\textit{Block3}}: $2s2x40d$     \\
\end{tabular}
\end{table}
\vspace{-4pt}

\noindent Note that \textbf{\textit{Block1}} is the building block of ResNeSt-50~\cite{Zhang2020ResNeStSN}.

We deliberately design such search space by two considerations.
\textbf{First}, these four candidate blocks have similar Params and FLOPs to avoid performance gain at the cost of model complexity since models with higher complexity often achieve higher accuracy. Thus, our search space is a nontrivial space to examine NAS's effectiveness. 
\textbf{Second}, our search space is similar with ResNet rather than the recent works \cite{guo2020single,wu2019fbnet,li2020block, Liu2018DARTSDA} since the experiments demonstrate that variants of ResNet are more efficient in practice even though the statics are in the opposite. As shown in Table \ref{tab:imagenet}, with the same top-1 accuracy on ImageNet, the latency of ResNeSt-50 surpasses EfficienNet-B3 \cite{Tan2019EfficientNetRM} by a margin of 14.5\% even though with $2.9\times$ more FLOPs. To further reduce the training consistency shift, we share the bottleneck's downsample operation among all candidate blocks in the same layer. The advantage of \textit{downsample-sharing} strategy will be illustrated in Section \ref{sec:ablation_study}.

Our $\Pi$-NAS is evaluated on ImageNet, a state-of-the-art classification dataset widely used in recent NAS methods \cite{guo2020single,wu2019fbnet,li2020block}. For the search procedure, we randomly pick out 50 images per class from the original 1.28M training set to build a 50k validation set, and the reset of images is used as a training set for supernet learning. All of our ImageNet results are tested on the original validation set.

\noindent\textbf{Training details.} We perform our $\Pi$-NAS in $3$ stages: $\Pi$-NAS learning, linear evaluation, and architecture search.

In $\Pi$-NAS learning, inspired by \cite{Chen2020ASF}, we use an augmentation strategy of random resize\&crop, color jitter, color drop, Gaussian blur, and horizontal flip. Besides, we employ a 2-layer MLP as the supernet head. The smoothing coefficient $\lambda$ of the \textit{mean teacher} in Eqn. \eqref{eq: mean teacher} is set to 0.999 in practice. The relative consistency loss is optimized by an SGD optimizer with a learning rate of 0.03, a momentum of 0.9, and a weight decay of $10^{-4}$. We adopt a cosine decay learning rate schedule to train for 100 epochs with a total batch size of 192 on 8 NVIDIA GTX 2080Ti GPUs.


As for linear evaluation, we fetch the optimized supernet-$\Pi$ model and replace the 2-layer MLP with a random initialized 1000-dimensional linear classifier. Only the linear classifier is trained on ImageNet for 100 epochs while the supernet's parameters $\mathcal{W}$ are frozen. At each training step, the linear classifier's inputs are obtained across stochastic paths from the supernet. Note that the batch statistics are used instead of tracked statistics in batch normalization (BN) layers to avoid inaccurate statistics across different sampled paths. Only random resize\&crop, horizontal flip are used for data augmentation. We train the classifier with a total batch size of 256 for 100 epochs using a cross-entropy loss and an SGD optimizer with an initial learning rate of 30, a momentum of 0.9, and a weight decay of 0. The learning rate decays by $0.1$ at 60 and 80 epochs.

In architecture search, the candidate architectures are evaluated separately with the top-1 accuracy on the 50k Imagenet validation set mentioned above. Again, to avoid the inaccurate batch statistics in BN, we pick out a further 50k images from the rest of the training set to recalculate the statistics for each optional path.
Then, we adopt a search algorithm, Action Space \cite{Wang2019SampleEfficientNA}, to seek candidates with the best performance with a maximum sample size of 1000.

\subsection{Experiments on ImageNet}\label{sec:imagenet experiments}
\noindent\textbf{Fast results of searched models.} As shown in Table \ref{tab:fast results}, we first evaluate the top 5 models searched by our $\Pi$-NAS as well as the ResNeSt-50 (\textit{Block1}) in a fast training setting. All the models are trained from scratch on the original ImageNet training set for 270 epochs with PyTorch-Encoding \cite{encoding2018} following the same setting of ResNeSt-50 except using a total batch size of 512 instead of 8192 due to the limit of GPU memory. Our models significantly outperform ResNeSt-50 by an average margin of 0.4\%, even with fewer parameters and FLOPs. In particular, all the searched top models achieve similar top-1 accuracy in supernet and training from scratch, respectively, which proves the effectiveness of our $\Pi$-NAS from another side.

\begin{table}
\caption{Image classification fast results on the validation set. (Acc@$S$: top-1 accuracy in supernet)}
\label{tab:fast results}
\small
\vspace{-16pt}
\begin{center}
    \setlength\tabcolsep{3pt}
    \begin{tabular}{l|c c|c|c c}
    \hline
    Model               &Params &FLOPs &Acc@$S$ &Acc@1 &Acc@5\\
    \hline\hline
    ResNeSt-50                &27.5M  &5.42G &64.6\%  &80.7\%  &95.3\% \\
    \hline\hline
    $\Pi$-NAS-$\alpha$ (ours)      &27.1M  &5.38G &65.0\%  &81.2\%  &95.4\% \\
    $\Pi$-NAS-$\beta$ (ours)         &27.2M  &5.39G &65.1\%  &81.2\%  &95.6\% \\
    $\Pi$-NAS-$\gamma$ (ours)        &27.0M  &5.30G &65.0\%  &81.1\%  &95.6\% \\
    $\Pi$-NAS-$\delta$ (ours)      &26.9M  &5.30G &65.0\%  &81.0\%  &95.4\% \\
    $\Pi$-NAS-$\epsilon$ (ours)        &26.9M  &5.42G &65.0\%  &81.0\%  &95.4\% \\
    \hline
    \end{tabular}
\vspace{-16pt}
\end{center}
\end{table}

\noindent\textbf{Comparison with the state-of-the-art models.} We select one of the searched models $\Pi$-NAS-$\alpha$ as our best model, denoted as $\Pi$-NAS-\textit{cls}, on ImageNet classification, considering a trade-off between performance and efficiency. We retrain ResNet-50 \cite{He2016DeepRL} (always undertrained in previous NAS works), ResNeSt-50 and our searched models on ImageNet under the same settings with an augmentation scheme, named AugMix \cite{Hendrycks2020AugMixAS}. For a fair comparison with the state-of-the-art NAS methods, we apply them on our search space $\mathcal{A}$. For SPOS~\cite{guo2020single} and FairNAS~\cite{Chu2019FairNA_arxiv}, we manipulate the same architecture search procedure as ours. For DNA~\cite{li2020block}, we select the candidate block with the minimum loss in each layer to build as its top model. For FBNetV2~\cite{Wan_2020_CVPR} and TuNAS~\cite{Bender_2020_CVPR}, we treat our search space as four possible channel decisions in each layer to apply the channel masking scheme. As we can see in Table \ref{tab:imagenet}, $\Pi$-NAS-\textit{cls} marks a new state-of-the-art top-1 accuracy 81.6\%, surpassing ResNeSt-50 by a large margin of 0.5\% in a similar computation complexity. By contrast, in our nontrivial search space, the previous NAS methods seem stuck at the local optima near ResNeSt-50, verifying the advantage of $\Pi$-NAS to reduce the supernet training consistency shift. Moreover, even though having more computation complexity, our $\Pi$-NAS-\textit{cls} achieves higher performance than EfficientNet-B3 \cite{Tan2019EfficientNetRM} with lower latency and less GPU memory in practice. Notably, the results in Table \ref{tab:imagenet} suggest that our $\Pi$-NAS-\textit{cls} not only achieves state-of-the-art performance but also runs at a fast speed indeed.

\begin{table}
\caption{Image classification results on the validation set. \textit{img/sec} and \textit{GPU} denote the inference speed and the GPU memory usage of the model performing inference, respectively, with a batch size of 128 on one NVIDIA GTX 2080Ti GPU. $\dagger$ Using crop size 300, otherwise 224. $\ast$ Searching for a bunch of sub-optimal solutions with close accuracy.}
\label{tab:imagenet}
\small
\vspace{-16pt}
\begin{center}
    \setlength\tabcolsep{1.2pt}
    \begin{tabular}{l|c c|c c|c}
    \hline
    Model           &Params &FLOPs &img/sec &GPU &Accuracy\\
    \hline\hline
    ResNet-50~\cite{He2016DeepRL} &25.6M  &4.12G &835.9 &2.55G &78.4  \\
    SENet-50~\cite{hu2018senet} &27.7M  &4.25G &-     &-     &78.9 \\
    SKNet-50~\cite{li2019selective} &27.5M  &4.47G &-     &-     &79.2 \\
    EfficientNet-B3$^\dagger$\cite{Tan2019EfficientNetRM} &12.2M  &1.88G &490.5 &9.25G &81.1  \\
    ResNeSt-50~\cite{Zhang2020ResNeStSN} &27.5M  &5.42G &561.6 &4.16G &81.1  \\
    \hline\hline
    \multicolumn{6}{l}{\textit{Searched Models on Our Search Space $\mathcal{A}$ from NAS Methods }}\\
    \hline
    SPOS~\cite{guo2020single} &27.1M  &5.43G &536.4 &4.12G &81.04$\pm$0.03 \\
    FairNAS~\cite{Chu2019FairNA_arxiv} &26.9M  &5.31G &541.7 &3.87G &81.05$\pm$0.06  \\
    DNA~\cite{li2020block} &26.8M  &5.41G &571.6 &3.71G &81.1*  \\
    FBNetV2~\cite{Wan_2020_CVPR} &26.8M  &5.29G &478.7 &3.89G &81.1*  \\
    TuNAS~\cite{Bender_2020_CVPR} &26.8M  &5.39G &554.8 &4.95G &81.1*  \\
    \textbf{\bm{$\Pi$}-NAS-\textit{cls} (ours)} &27.1M  &5.38G &556.8 &4.07G &\textbf{81.6} \\
    \hline
    \end{tabular}
\vspace{-22pt}
\end{center}
\end{table}

\noindent\textbf{Model ranking.} As discussed in Section \ref{sec: intro}, a strong ranking correlation between candidates' actual and predicted performance in the supernet is essential to the effectiveness of NAS. Here, we compare our ranking correlation with DNA~\cite{li2020block} and SPOS~\cite{guo2020single}. We use the top 5 architectures in Table \ref{tab:fast results} and randomly sample other eight architectures from the search space and train them in a fast setting described above to obtain their top-1 accuracy training from scratch, then fetch their predicted performances in the supernet of each method to compute the ranking correlations. The second row of Table~\ref{tab:ranking_corr} suggests the advanced effectiveness of $\Pi$-NAS as it predicts the model's performance much more correctly. As analyzed in Section~\ref{sec:dilemma}, this is due to the training consistency shift problem, which will be further discussed in Section \ref{sec:ablation_study}.

\begingroup
\setlength{\tabcolsep}{3pt} 
\begin{table}
\caption{Ranking correlations (in Kendall’s Tau metric) of diverse NAS methods in our search space.}
\label{tab:ranking_corr}
\small
\vspace{-16pt}
\begin{center}
    \setlength\tabcolsep{2.3pt}
  \begin{tabular}{l|c|c|c|c|c|c}
    \hline
    Method          &\textbf{Ours} &DNA &SPOS &FairNAS &FBNetV2 &TuNAS\\
    \hline
    Classification  &\textbf{0.79}      &0.45      &0.19   &0.36   &0.32   &0.14\\
    \hline
    Instance seg.  &\textbf{0.51}      &0.38        &0.18   &-  &-  &-\\
    \hline
    \end{tabular}
\end{center}
\vspace{-10pt}
\end{table}
\endgroup

\begin{figure}
    \centering
    \includegraphics[width=.9\linewidth]{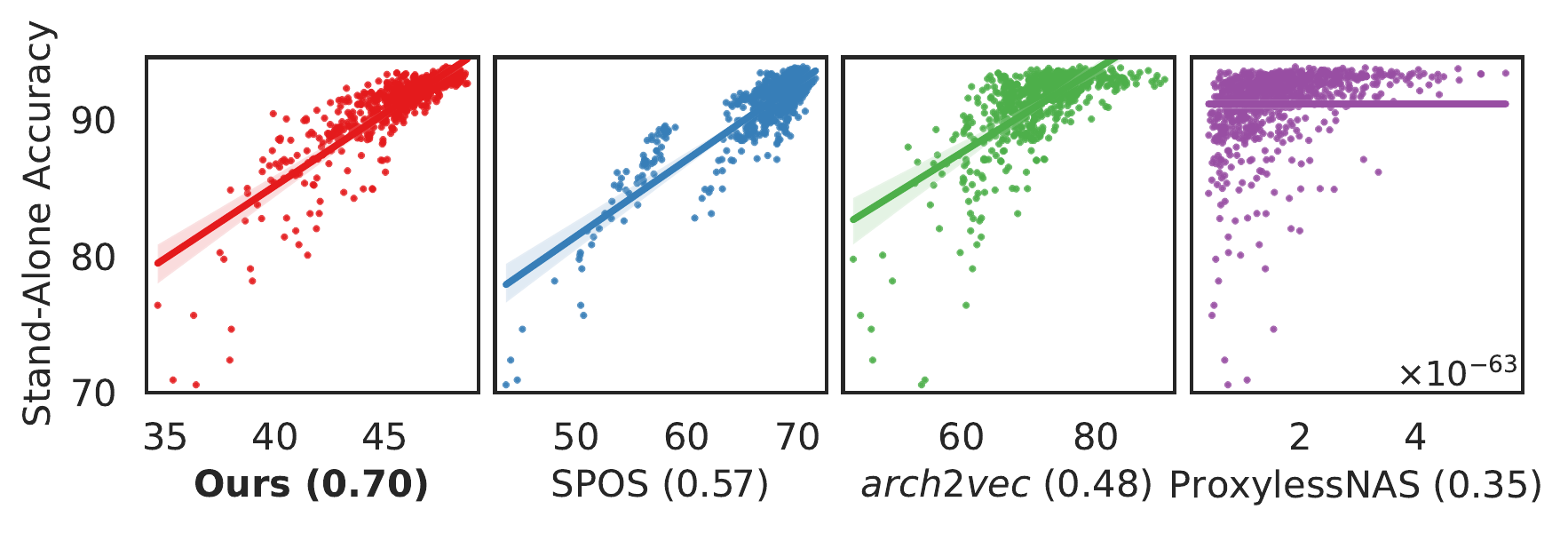}
    \vspace{-4pt}
    \caption{Ranking correlations on 792 architectures on NAS-Bench-201~\cite{dong2020nasbench201} on CIFAR-10 without \textit{skip connection} and \textit{zero} operations compared to SPOS~\cite{guo2020single}, \textit{arch2vec}~\cite{yan2020arch} and ProxylessNAS~\cite{cai2018proxylessnas}.}
    \label{fig:nasbench_correlation}
\end{figure}

\begin{table}
\caption{Results on NAS-Bench-201 on CIFAR-10.}
\label{tab:nasbench}
\small
\vspace{-14pt}
\scriptsize
\begin{center}
    \setlength\tabcolsep{1pt}
    \begin{tabular}{l|c|c|c|c|c|c}
    \hline
    Method & Ours & SPOS & \textit{arch2vec} & ProxylessNAS & WPL & GDAS-NSAS \\
    \hline
    Test(\%) & \textbf{93.83$\pm$0} & 93.57$\pm$0 & 92.53$\pm$0.32 & 92.08$\pm$0.03 & 90.92$\pm$0.11 & 93.55$\pm$0.16 \\
    \hline
    \end{tabular}
\vspace{-10pt}
\end{center}
\end{table}

\begin{table}
\caption{Instance segmentation results with Mask-RCNN \cite{He2017MaskR} on the COCO 2017 validation set.}
\label{tab:coco}
\small
\vspace{-12pt}
\begin{center}
    \setlength\tabcolsep{12pt}
    \begin{tabular}{l|cc}
    \hline
    Model         & AP$^{Box}$ &AP$^{Mask}$\\
    \hline\hline
    ResNet-50 \cite{He2016DeepRL}         &39.93$\pm$0.04    &35.99$\pm$0.06\\
    ResNeSt-50 \cite{Zhang2020ResNeStSN}  &42.81$\pm$0.02    &38.14$\pm$0.01\\
    \hline\hline
    $\Pi$-NAS-\textit{cls} (ours)    &{43.72}           &{39.13}\\
    \textbf{\bm{$\Pi$}-NAS-\textit{trans} (ours)}  &\textbf{44.11$\pm$0.04}    &\textbf{39.48$\pm$0.02}\\
    \hline
    \end{tabular}
\vspace{-12pt}
\end{center}
\end{table}

\begin{table}
\caption{Semantic segmentation results with DeeplabV3 \cite{Chen2017RethinkingAC} on the validation set of ADE20K and Cityscapes.}
\vspace{-12pt}
\label{tab:ade_and_citys}
\small
\begin{center}
    \setlength\tabcolsep{3.8pt}
    \begin{tabular}{l|cc|c}
    \hline
    \multirow{2}{*}{Model} & \multicolumn{2}{c|}{ADE20K} & Cityscapes\\
                           &pixAcc  &mIoU  &mIoU\\
    \hline\hline
    ResNet-50 \cite{He2016DeepRL}        &80.66$\pm$0.27  &42.74$\pm$0.64  &78.42$\pm$0.30\\
    ResNeSt-50 \cite{Zhang2020ResNeStSN} &81.22$\pm$0.05  &45.18$\pm$0.06  &80.08$\pm$0.20\\
    \hline\hline
    \textbf{\bm{$\Pi$}-NAS-\textit{trans} (ours)} &\textbf{81.31$\pm$0.04} &\textbf{45.49$\pm$0.02}  &\textbf{80.40$\pm$0.30}\\
    \hline
    \end{tabular}
\vspace{-16pt}
\end{center}
\end{table}

\begin{figure}
    \centering
    \includegraphics[width=1\linewidth]{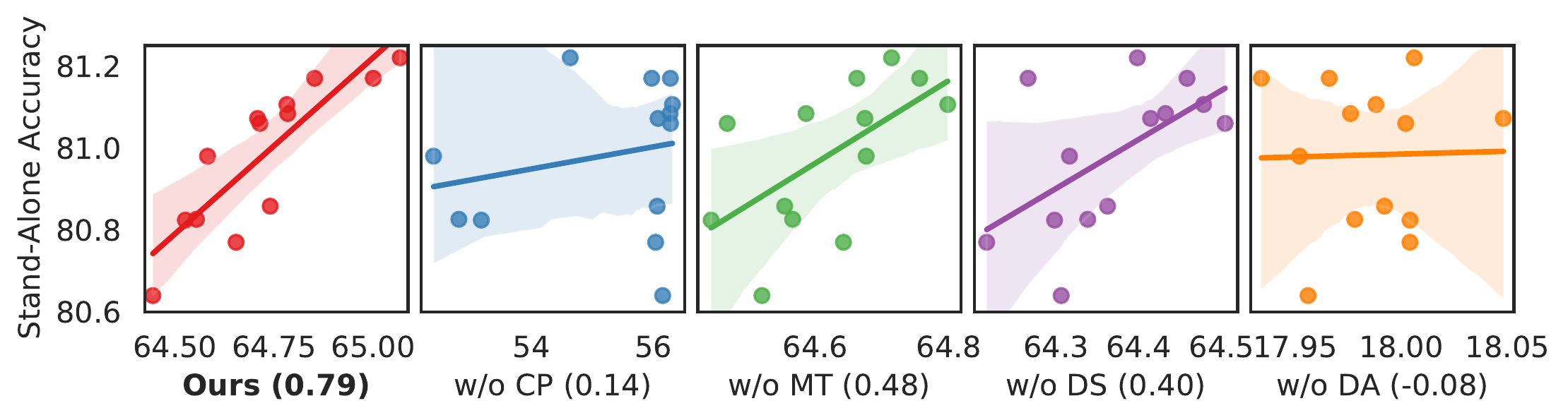}
    \caption{Ranking correlations of diverse NAS methods. Values within the parentheses are their corresponding Kendall’s Tau. Our $\Pi$-NAS gains the best correlation, indicating that $\Pi$-NAS effectively reduces the training consistency shift.(CP: \textit{cross-path learning}; MT: \textit{mean teacher}; DS: \textit{downsample-sharing}; DA: \textit{learning different augmented views of the same image})}
    \label{fig:ranking_correlation}
\vspace{-11pt}
\end{figure}

\begin{figure*}
    \centering
    \includegraphics[width=1\linewidth]{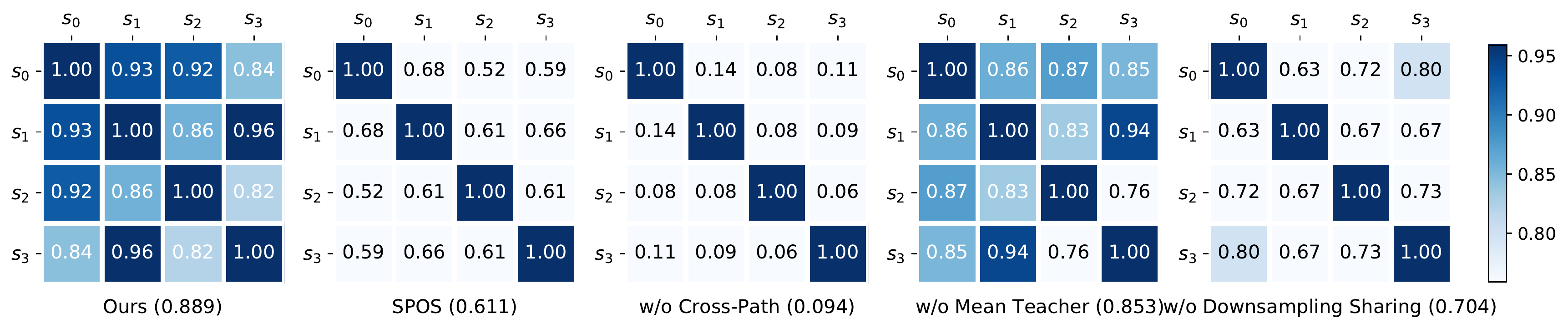}
    \vspace{-12pt}
    \caption{Cosine similarity matrices of the outputs from the last layer of $4$ paths in each supernet variant. The value within the parentheses is the average after stripping out the diagonal. Our $\Pi$-NAS achieves the overall highest feature similarity, intuitively explaining its effectiveness to reduce the supernet’s training consistency shift.}
    \label{fig:supernet_cosine_similarity}
\vspace{-12pt}
\end{figure*}

\subsection{Experimenet on NAS-Bench-201 Benchmarks}
We additionally validate our ${\Pi}$-NAS on a popular cell-based search space, NAS-Bench-201~\cite{dong2020nasbench201}, on CIFAR-10 dataset. This search space is represented as a DAG, where each edge is associated to an operation with 5 options: \textit{zero, skip connection, $1\times1$ convolution, $3\times3$ convolution} and \textit{$3\times3$ average pooling}. This DAG has 4 nodes, where each node represents the sum of feature maps transformed through the edges pointing to this node. For the sake of simplicity, though we train the supernet involving all 5 operations, we predict the performances of all 792 architectures without \textit{zero} and \textit{skip connection} operations to measure the ranking correlation to their ground-truth performances. As shown in Figure \ref{fig:nasbench_correlation} and Table \ref{tab:nasbench}, our method significantly outperforms SPOS~\cite{guo2020single}, \textit{arch2vec}~\cite{yan2020arch} (an unsupervised NAS method), ProxylessNAS~\cite{cai2018proxylessnas} (a differentiable method), WPL~\cite{benyahia2019overcoming} (a different solution to address parameter shift) and GDAS-NASA~\cite{zhang2020one} by a clear margin, verifying our method's effectiveness and compatibility.

\newcommand{\apbbox}[1]{AP$^\text{bb}_\text{#1}$}
\newcommand{\apmask}[1]{AP$^\text{mk}_\text{#1}$}

\subsection{Experiments on Transfer Learning}\label{sec:transfer learning}
\noindent\textbf{Instance segmentation results.}
To explore the transferability of our $\Pi$-NAS models, we first evaluate them on a widely used transfer learning task, instance segmentation, which simultaneously solves the problem of object detection and semantic segmentation. We train the Mask-RCNN \cite{He2017MaskR} on COCO-2017 with our searched models as its backbone following the instructions of \cite{Zhang2020ResNeStSN,Wang2020Grammatically_kdd}. Rather than one model, we evaluate all the 13 architectures (used in \ref{sec:imagenet experiments} \textit{Model ranking}) with pretrain models on ImageNet. Also, we study the ranking correlation by averaging the bounding box mAP (\apbbox{}) and mask mAP (\apmask{}) as the actual performance. As shown in the third row of Table~\ref{tab:ranking_corr}, the effectiveness of our $\Pi$-NAS stays superior, which indicates that our approach can search for architectures that are more transferrable and universal. Note that we choose the architecture with the best performance as a transferable model, $\Pi$-NAS-\textit{trans} (\textit{a.k.a.} $\Pi$-NAS-$\gamma$, one of our top 5 searched architectures), for the transfer learning. Table \ref{tab:coco} shows that both of $\Pi$-NAS-\textit{trans} and $\Pi$-NAS-\textit{cls} outperform ResNeSt-50 by a significant margin (0.91\% and 1.30\% in \apbbox{}).

\noindent\textbf{Semantic segmentation results.} 
We further transfer $\Pi$-NAS-\textit{trans} to the downstream task of semantic segmentation on ADE20K and Cityscapes datasets. We train DeeplabV3\cite{Chen2017RethinkingAC} with the implementation of PyTorch-Encoding and the settings from\cite{Zhang2020ResNeStSN}. For the ADE20K dataset, we train the model for 120 epochs with a base image size of 520 and cropped image size of 480. As for the Cityscapes dataset, the model is trained for 240 epochs; the base image size is 2048; the cropped image size is 768. We also follow \cite{Zhang2020ResNeStSN} to use multi-scale evaluation with ﬂipping. Results are shown in Table \ref{tab:ade_and_citys}, both of which demonstrate the advantage of our $\Pi$-NAS-\textit{trans}.

\subsection{Ablation Study}\label{sec:ablation_study}

\noindent\textbf{Effectiveness of components.} To evaluate the impact of our $\Pi$-NAS separately, we first distinguish it from SPOS by \textit{cross-path learning}, \textit{mean teacher} and \textit{downsample-sharing}. As shown in Table \ref{tab:components} and Figure \ref{fig:ranking_correlation}, we test the combination methods with Kendall's Tau as a ranking correlation between their models' predicted and actual performance. Adopting the same testing scheme in Section \ref{sec:imagenet experiments}, we apply each method on their supernets' training and then evaluate the 13 architectures (used in \ref{sec:imagenet experiments} \textit{Model ranking}). As we can see, the Supernet-$\Pi$ model reduces by 0.31 without the mean-teacher, which indicates \textit{mean teacher} plays a role in high ranking correlation. The most notable thing is that without \textit{cross-path learning}, the method lost its effectiveness as SPOS. Obviously, \textit{cross-path learning} is the essential component in our $\Pi$-NAS. \textit{Downsample-sharing} also shows its strength in predicting accurate performance for candidate architectures with a 0.39 improvement. Note that when we try to perform $\Pi$-NAS without \textit{nontrivial mean teacher}, the supernet converged quickly to a state that outputs all zeros, which disables the distinguishing ability of the model (see Table \ref{tab:components}).

\begin{table}
\caption{Effectiveness of each component of our $\Pi$-NAS. (CP: \textit{cross-path learning}; MT: \textit{mean teacher}; DS: \textit{downsample-sharing})}
\scriptsize
\vspace{-14pt}
\begin{center}
    \begin{tabular}{l|c|c|c|c|c}
    \hline
    Method      &CP      &MT      &DS     & nontrivial &Kendall's Tau\\
    \hline\hline
    SPOS~\cite{guo2020single}   &               &         &   \checkmark    &           &0.19\\
    S-$\Pi$ model               &\checkmark     &               &\checkmark &\checkmark &0.48\\
    Ours w/o CP                 &               &\checkmark     &\checkmark & \checkmark &0.14\\
    Ours w/o DS                 &\checkmark     &\checkmark     &         &\checkmark  &0.40\\
    Ours w/o nontrival                       & \checkmark    &\checkmark    & \checkmark &     &collision\\
    \textbf{Ours}                        &\checkmark     &\checkmark    & \checkmark & \checkmark & \textbf{0.79}\\
    \hline
    \end{tabular}
    \label{tab:components}
\end{center}
\vspace{-20pt}
\end{table}

\noindent\textbf{Feature consistency and ranking correlation.} As analyzed in Section \ref{sec:method}, \textit{training consistency shift} damages the ranking correlation of NAS. To further demonstrate this statement, we explore and visualize the feature similarity from the last layer across paths. For example, we randomly sample 4 architectures except the last layer are \textit{Block0}, \textit{Block1}, \textit{Block2} and \textit{Block3} respectively, which are denoted as $s_0$, $s_1$, $s_2$ and $s_3$. Then we evaluate the feature cosine similarity between each pair of them. Figure \ref{fig:supernet_cosine_similarity} shows the embedding feature similarity of different methods. By correlating Figure \ref{fig:ranking_correlation}, we found that a high feature consistency lead to a strong ranking correlation of supernet, which demonstrates convincingly our motivation. Notably, Figure \ref{fig:supernet_cosine_similarity} also proves our $\Pi$-NAS indeed reduces \textit{supernet training consistency shift}, especially for \textit{cross-path learning}.


\vspace{-4pt}
\section{Conclusion}
This paper recognizes the importance of architecture ranking in NAS and attributes the ranking correlation problem to the supernet training consistency shift, including feature shift an parameter shift. To address these two shifts, we propose a nontrivial supernet-$\Pi$ model, \ie, $\Pi$-NAS. Specifically, we propose a supernet-$\Pi$ model with cross-path learning to reduce feature shift and a nontrivial mean teacher to cope with parameter shift. Notably, our $\Pi$-NAS can search for more transferable and universal architectures than supervised NAS. Extensive experiments on many tasks demonstrate the search effectiveness and universality of our $\Pi$-NAS compared to the NAS counterparts.

\vspace{-7pt}
\section*{Acknowledgement}
\vspace{-7pt}
This work was supported in part by National Key R\&D Program of China under Grant No.2020AAA0109700, National Natural Science Foundation of China (U19A2073 and 61976233), Guangdong Province Basic and Applied Basic Research (2019B1515120039), Guangdong Outstanding Youth Fund (2021B1515020061), Shenzhen Fundamental Research Program (RCYX20200714114642083, JCYJ20190807154211365), Zhejiang Lab’s Open Fund (2020AA3AB14) and CSIG Young Fellow Support Fund.

{\small

\bibliographystyle{ieee_fullname}

\begin{thebibliography}{10}\itemsep=-1pt

\bibitem{akimoto2019adaptive}
Youhei Akimoto, Shinichi Shirakawa, Nozomu Yoshinari, Kento Uchida, Shota
  Saito, and Kouhei Nishida.
\newblock Adaptive stochastic natural gradient method for one-shot neural
  architecture search.
\newblock In {\em ICML}, pages 171--180, 2019.

\bibitem{bachman2014learning}
Philip Bachman, Ouais Alsharif, and Doina Precup.
\newblock Learning with pseudo-ensembles.
\newblock In {\em NeurIPS}, pages 3365--3373, 2014.

\bibitem{baker2016designing}
Bowen Baker, Otkrist Gupta, Nikhil Naik, and Ramesh Raskar.
\newblock Designing neural network architectures using reinforcement learning.
\newblock {\em arXiv preprint arXiv:1611.02167}, 2016.

\bibitem{bender2018understanding}
Gabriel Bender, Pieter{-}Jan Kindermans, Barret Zoph, Vijay Vasudevan, and
  Quoc~V. Le.
\newblock Understanding and simplifying one-shot architecture search.
\newblock In {\em ICML}, pages 549--558, 2018.

\bibitem{Bender_2020_CVPR}
Gabriel Bender, Hanxiao Liu, Bo Chen, Grace Chu, Shuyang Cheng, Pieter-Jan
  Kindermans, and Quoc~V. Le.
\newblock Can weight sharing outperform random architecture search? an
  investigation with tunas.
\newblock In {\em CVPR}, June 2020.

\bibitem{benyahia2019overcoming}
Yassine Benyahia, Kaicheng Yu, Kamil~Bennani Smires, Martin Jaggi, Anthony~C
  Davison, Mathieu Salzmann, and Claudiu Musat.
\newblock Overcoming multi-model forgetting.
\newblock In {\em International Conference on Machine Learning}, pages
  594--603. PMLR, 2019.

\bibitem{berthelot2019mixmatch}
David Berthelot, Nicholas Carlini, Ian Goodfellow, Nicolas Papernot, Avital
  Oliver, and Colin~A Raffel.
\newblock Mixmatch: A holistic approach to semi-supervised learning.
\newblock In {\em NeurIPS}, pages 5049--5059, 2019.

\bibitem{brock2017smash}
Andrew Brock, Theodore Lim, James~M. Ritchie, and Nick Weston.
\newblock {SMASH:} one-shot model architecture search through hypernetworks.
\newblock In {\em ICLR}, 2018.

\bibitem{cai2018proxylessnas}
Han Cai, Ligeng Zhu, and Song Han.
\newblock Proxyless{NAS}: Direct neural architecture search on target task and
  hardware.
\newblock In {\em ICLR}, 2019.

\bibitem{Chen2017RethinkingAC}
Liang-Chieh Chen, G. Papandreou, Florian Schroff, and H. Adam.
\newblock Rethinking atrous convolution for semantic image segmentation.
\newblock {\em ArXiv}, abs/1706.05587, 2017.

\bibitem{chen2020simple}
Ting Chen, Simon Kornblith, Mohammad Norouzi, and Geoffrey Hinton.
\newblock A simple framework for contrastive learning of visual
  representations.
\newblock {\em arXiv preprint arXiv:2002.05709}, 2020.

\bibitem{Chen2020ASF}
Ting Chen, Simon Kornblith, Mohammad Norouzi, and Geoffrey~E. Hinton.
\newblock A simple framework for contrastive learning of visual
  representations.
\newblock {\em ArXiv}, abs/2002.05709, 2020.

\bibitem{chen2020improved}
Xinlei Chen, Haoqi Fan, Ross Girshick, and Kaiming He.
\newblock Improved baselines with momentum contrastive learning.
\newblock {\em arXiv preprint arXiv:2003.04297}, 2020.

\bibitem{chen2018reinforced}
Yukang Chen, Gaofeng Meng, Qian Zhang, Shiming Xiang, Chang Huang, Lisen Mu,
  and Xinggang Wang.
\newblock {RENAS:} reinforced evolutionary neural architecture search.
\newblock In {\em CVPR}, pages 4787--4796, 2019.

\bibitem{chu2019scarletnas}
Xiangxiang Chu, Bo Zhang, Jixiang Li, Qingyuan Li, and Ruijun Xu.
\newblock Scarletnas: Bridging the gap between scalability and fairness in
  neural architecture search.
\newblock {\em arXiv preprint arXiv:1908.06022}, 2019.

\bibitem{Chu2019FairNA_arxiv}
Xiangxiang Chu, Bo Zhang, Ruijun Xu, and Jixiang Li.
\newblock Fairnas: Rethinking evaluation fairness of weight sharing neural
  architecture search.
\newblock {\em CoRR}, abs/1907.01845, 2019.

\bibitem{dong2019searching}
Xuanyi Dong and Yi Yang.
\newblock Searching for a robust neural architecture in four {GPU} hours.
\newblock In {\em CVPR}, pages 1761--1770, 2019.

\bibitem{dong2020nasbench201}
Xuanyi Dong and Yi Yang.
\newblock Nas-bench-201: Extending the scope of reproducible neural
  architecture search.
\newblock In {\em ICLR}, 2020.

\bibitem{grill2020bootstrap}
Jean-Bastien Grill, Florian Strub, Florent Altch{\'e}, Corentin Tallec,
  Pierre~H Richemond, Elena Buchatskaya, Carl Doersch, Bernardo~Avila Pires,
  Zhaohan~Daniel Guo, Mohammad~Gheshlaghi Azar, et~al.
\newblock Bootstrap your own latent: A new approach to self-supervised
  learning.
\newblock {\em arXiv preprint arXiv:2006.07733}, 2020.

\bibitem{guo2020single}
Zichao Guo, Xiangyu Zhang, Haoyuan Mu, Wen Heng, Zechun Liu, Yichen Wei, and
  Jian Sun.
\newblock Single path one-shot neural architecture search with uniform
  sampling.
\newblock In {\em ECCV}. Springer, 2020.

\bibitem{hadsell2006dimensionality}
Raia Hadsell, Sumit Chopra, and Yann LeCun.
\newblock Dimensionality reduction by learning an invariant mapping.
\newblock In {\em CVPR}. IEEE, 2006.

\bibitem{he2020momentum}
Kaiming He, Haoqi Fan, Yuxin Wu, Saining Xie, and Ross Girshick.
\newblock Momentum contrast for unsupervised visual representation learning.
\newblock In {\em CVPR}, 2020.

\bibitem{He2017MaskR}
Kaiming He, Georgia Gkioxari, Piotr Doll{\'a}r, and Ross~B. Girshick.
\newblock Mask r-cnn.
\newblock {\em ICCV}, 2017.

\bibitem{He2016DeepRL}
Kaiming He, X. Zhang, Shaoqing Ren, and Jian Sun.
\newblock Deep residual learning for image recognition.
\newblock {\em CVPR}, 2016.

\bibitem{Hendrycks2020AugMixAS}
Dan Hendrycks, Norman Mu, E.~D. Cubuk, Barret Zoph, J. Gilmer, and Balaji
  Lakshminarayanan.
\newblock Augmix: A simple data processing method to improve robustness and
  uncertainty.
\newblock {\em ArXiv}, abs/1912.02781, 2020.

\bibitem{hjelm2018learning}
R~Devon Hjelm, Alex Fedorov, Samuel Lavoie-Marchildon, Karan Grewal, Phil
  Bachman, Adam Trischler, and Yoshua Bengio.
\newblock Learning deep representations by mutual information estimation and
  maximization.
\newblock {\em arXiv preprint arXiv:1808.06670}, 2018.

\bibitem{hu2018senet}
Jie Hu, Li Shen, and Gang Sun.
\newblock Squeeze-and-excitation networks.
\newblock In {\em CVPR}, 2018.

\bibitem{kolesnikov2019revisiting}
Alexander Kolesnikov, Xiaohua Zhai, and Lucas Beyer.
\newblock Revisiting self-supervised visual representation learning.
\newblock In {\em CVPR}, pages 1920--1929, 2019.

\bibitem{laine2016temporal}
Samuli Laine and Timo Aila.
\newblock Temporal ensembling for semi-supervised learning.
\newblock {\em arXiv preprint arXiv:1610.02242}, 2016.

\bibitem{li2020block}
Changlin Li, Jiefeng Peng, Liuchun Yuan, Guangrun Wang, Xiaodan Liang, Liang
  Lin, and Xiaojun Chang.
\newblock Block-wisely supervised neural architecture search with knowledge
  distillation.
\newblock In {\em CVPR}, 2020.

\bibitem{li2021bossnas}
Changlin Li, Tao Tang, Guangrun Wang, Jiefeng Peng, Bing Wang, Xiaodan Liang,
  and Xiaojun Chang.
\newblock {B}oss{NAS}: Exploring hybrid {CNN}-transformers with block-wisely
  self-supervised neural architecture search.
\newblock In {\em ICCV}, 2021.

\bibitem{LiWWLLC21}
Changlin Li, Guangrun Wang, Bing Wang, Xiaodan Liang, Zhihui Li, and Xiaojun
  Chang.
\newblock {Dynamic Slimmable Network}.
\newblock In {\em CVPR}, 2021.

\bibitem{li2020improving}
Xiang Li, Chen Lin, Chuming Li, Ming Sun, Wei Wu, Junjie Yan, and Wanli Ouyang.
\newblock Improving one-shot nas by suppressing the posterior fading.
\newblock In {\em CVPR}, 2020.

\bibitem{li2019selective}
Xiang Li, Wenhai Wang, Xiaolin Hu, and Jian Yang.
\newblock Selective kernel networks.
\newblock In {\em CVPR}, 2019.

\bibitem{liu2020labels}
Chenxi Liu, Piotr Doll{\'a}r, Kaiming He, Ross Girshick, Alan Yuille, and
  Saining Xie.
\newblock Are labels necessary for neural architecture search?
\newblock {\em arXiv preprint arXiv:2003.12056}, 2020.

\bibitem{Liu2018DARTSDA}
Hanxiao Liu, Karen Simonyan, and Yiming Yang.
\newblock {DARTS:} differentiable architecture search.
\newblock In {\em ICLR}, 2019.

\bibitem{negrinho2017deeparchitect}
Renato Negrinho and Geoffrey~J. Gordon.
\newblock Deeparchitect: Automatically designing and training deep
  architectures.
\newblock {\em CoRR}, abs/1704.08792, 2017.

\bibitem{oliver2018realistic}
Avital Oliver, Augustus Odena, Colin~A Raffel, Ekin~Dogus Cubuk, and Ian
  Goodfellow.
\newblock Realistic evaluation of deep semi-supervised learning algorithms.
\newblock In {\em NeurIPS}, pages 3235--3246, 2018.

\bibitem{oord2018representation}
Aaron van~den Oord, Yazhe Li, and Oriol Vinyals.
\newblock Representation learning with contrastive predictive coding.
\newblock {\em arXiv preprint arXiv:1807.03748}, 2018.

\bibitem{real2017large}
Esteban Real, Sherry Moore, Andrew Selle, Saurabh Saxena, Yutaka~Leon Suematsu,
  Jie Tan, Quoc Le, and Alex Kurakin.
\newblock Large-scale evolution of image classifiers.
\newblock {\em arXiv preprint arXiv:1703.01041}, 2017.

\bibitem{sajjadi2016regularization}
Mehdi Sajjadi, Mehran Javanmardi, and Tolga Tasdizen.
\newblock Regularization with stochastic transformations and perturbations for
  deep semi-supervised learning.
\newblock In {\em NeurIPS}, pages 1163--1171, 2016.

\bibitem{tan2019mnasnet}
Mingxing Tan, Bo Chen, Ruoming Pang, Vijay Vasudevan, Mark Sandler, Andrew
  Howard, and Quoc~V Le.
\newblock Mnasnet: Platform-aware neural architecture search for mobile.
\newblock In {\em CVPR}, pages 2820--2828, 2019.

\bibitem{Tan2019EfficientNetRM}
M. Tan and Quoc~V. Le.
\newblock Efficientnet: Rethinking model scaling for convolutional neural
  networks.
\newblock {\em ArXiv}, abs/1905.11946, 2019.

\bibitem{tarvainen2017mean}
Antti Tarvainen and Harri Valpola.
\newblock Mean teachers are better role models: Weight-averaged consistency
  targets improve semi-supervised deep learning results.
\newblock In {\em NeurIPS}, 2017.

\bibitem{tian2019contrastive}
Yonglong Tian, Dilip Krishnan, and Phillip Isola.
\newblock Contrastive multiview coding.
\newblock {\em arXiv preprint arXiv:1906.05849}, 2019.

\bibitem{Wan_2020_CVPR}
Alvin Wan, Xiaoliang Dai, Peizhao Zhang, Zijian He, Yuandong Tian, Saining Xie,
  Bichen Wu, Matthew Yu, Tao Xu, Kan Chen, Peter Vajda, and Joseph~E. Gonzalez.
\newblock Fbnetv2: Differentiable neural architecture search for spatial and
  channel dimensions.
\newblock In {\em CVPR}, June 2020.

\bibitem{wang2020Smoothing_cvpr}
Guangcong Wang, Jian-Huang Lai, Wenqi Liang, and Guangrun Wang.
\newblock Smoothing adversarial domain attack and p-memory reconsolidation for
  cross-domain person re-identification.
\newblock In {\em Proceedings of the IEEE/CVF Conference on Computer Vision and
  Pattern Recognition (CVPR)}, June 2020.

\bibitem{wang2021Joint_tnnls}
Guangrun Wang, Liang Lin, Rongcong Chen, Guangcong Wang, and Jiqi Zhang.
\newblock Joint learning of neural transfer and architecture adaptation for
  image recognition.
\newblock {\em IEEE Transactions on Neural Networks and Learning Systems
  (T-NNLS)}, 2021.

\bibitem{Wang2020Grammatically_kdd}
Guangrun Wang, Guangcong Wang, Keze Wang, Xiaodan Liang, and Liang Lin.
\newblock Grammatically recognizing images with tree convolution.
\newblock In Rajesh Gupta, Yan Liu, Jiliang Tang, and B.~Aditya Prakash,
  editors, {\em {KDD} '20: The 26th {ACM} {SIGKDD} Conference on Knowledge
  Discovery and Data Mining, Virtual Event, CA, USA, August 23-27, 2020}, pages
  903--912. {ACM}, 2020.

\bibitem{wang2020weakly_tnnls}
Guangrun Wang, Guangcong Wang, Xujie Zhang, Jianhuang Lai, Zhengtao Yu, and
  Liang Lin.
\newblock Weakly supervised person re-id: Differentiable graphical learning and
  a new benchmark.
\newblock {\em IEEE Transactions on Neural Networks and Learning Systems},
  32(5):2142--2156, 2020.

\bibitem{wang2021Solving_iccv}
Guangrun Wang, Keze Wang, Guangcong Wang, Philip H.~S. Torr, and Liang Lin.
\newblock Solving inefficiency of self-supervised representation learning.
\newblock 2021.

\bibitem{wang2017deep_iccv}
Guangcong Wang, Xiaohua Xie, Jianhuang Lai, and Jiaxuan Zhuo.
\newblock Deep growing learning.
\newblock In {\em Proceedings of the IEEE International Conference on Computer
  Vision}, pages 2812--2820, 2017.

\bibitem{Wang2019SampleEfficientNA}
Linnan Wang, Saining Xie, Teng Li, Rodrigo Fonseca, and Yuandong Tian.
\newblock Sample-efficient neural architecture search by learning action space.
\newblock {\em ArXiv}, abs/1906.06832, 2019.

\bibitem{wu2019fbnet}
Bichen Wu, Xiaoliang Dai, Peizhao Zhang, Yanghan Wang, Fei Sun, Yiming Wu,
  Yuandong Tian, Peter Vajda, Yangqing Jia, and Kurt Keutzer.
\newblock Fbnet: Hardware-aware efficient convnet design via differentiable
  neural architecture search.
\newblock In {\em CVPR}, 2019.

\bibitem{wu2018unsupervised}
Zhirong Wu, Yuanjun Xiong, Stella~X Yu, and Dahua Lin.
\newblock Unsupervised feature learning via non-parametric instance
  discrimination.
\newblock In {\em CVPR}, pages 3733--3742, 2018.

\bibitem{xie2019unsupervised}
Qizhe Xie, Zihang Dai, Eduard Hovy, Minh-Thang Luong, and Quoc~V Le.
\newblock Unsupervised data augmentation for consistency training.
\newblock {\em arXiv preprint arXiv:1904.12848}, 2019.

\bibitem{xie2020self}
Qizhe Xie, Minh-Thang Luong, Eduard Hovy, and Quoc~V Le.
\newblock Self-training with noisy student improves imagenet classification.
\newblock In {\em CVPR}, pages 10687--10698, 2020.

\bibitem{yan2020does}
Shen Yan, Yu Zheng, Wei Ao, Xiao Zeng, and Mi Zhang.
\newblock Does unsupervised architecture representation learning help neural
  architecture search?
\newblock {\em NeurIPS}, 33, 2020.

\bibitem{yan2020arch}
Shen Yan, Yu Zheng, Wei Ao, Xiao Zeng, and Mi Zhang.
\newblock Does unsupervised architecture representation learning help neural
  architecture search?
\newblock In {\em NeurIPS}, 2020.

\bibitem{zhai2019s4l}
Xiaohua Zhai, Avital Oliver, Alexander Kolesnikov, and Lucas Beyer.
\newblock S4l: Self-supervised semi-supervised learning.
\newblock In {\em ICCV}, pages 1476--1485, 2019.

\bibitem{encoding2018}
Hang Zhang.
\newblock {PyTorch-Encoding}.
\newblock \url{https://github.com/zhanghang1989/PyTorch-Encoding}, 2018.

\bibitem{Zhang2020ResNeStSN}
Hang Zhang, Chongruo Wu, Zhongyue Zhang, Yi Zhu, Zhi-Li Zhang, Haibin Lin, Yu e
  Sun, Tong He, Jonas Mueller, R. Manmatha, M. Li, and Alex Smola.
\newblock Resnest: Split-attention networks.
\newblock {\em ArXiv}, abs/2004.08955, 2020.

\bibitem{ZhangLPCGS20}
Miao Zhang, Huiqi Li, Shirui Pan, Xiaojun Chang, Zongyuan Ge, and Steven~W. Su.
\newblock Differentiable neural architecture search in equivalent space with
  exploration enhancement.
\newblock In {\em {NeurIPS}}, 2020.

\bibitem{zhang2020overcoming}
Miao Zhang, Huiqi Li, Shirui Pan, Xiaojun Chang, and Steven Su.
\newblock Overcoming multi-model forgetting in one-shot nas with diversity
  maximization.
\newblock In {\em CVPR}, pages 7809--7818, 2020.

\bibitem{zhang2020one}
Miao Zhang, Huiqi Li, Shirui Pan, Xiaojun Chang, Chuan Zhou, Zongyuan Ge, and
  Steven~W Su.
\newblock One-shot neural architecture search: Maximising diversity to overcome
  catastrophic forgetting.
\newblock {\em IEEE Annals of the History of Computing}, 2020.

\bibitem{zhang2021randomlabelnas}
Xuanyang Zhang, Pengfei Hou, Xiangyu Zhang, and Jian Sun.
\newblock Neural architecture search with random labels.
\newblock In {\em CVPR}, 2021.

\bibitem{zhong2018practical}
Zhao Zhong, Junjie Yan, Wei Wu, Jing Shao, and Cheng{-}Lin Liu.
\newblock Practical block-wise neural network architecture generation.
\newblock In {\em CVPR}, pages 2423--2432, 2018.

\bibitem{zhuang2019local}
Chengxu Zhuang, Alex~Lin Zhai, and Daniel Yamins.
\newblock Local aggregation for unsupervised learning of visual embeddings.
\newblock In {\em ICCV}, pages 6002--6012, 2019.

\bibitem{zoph2016neural}
Barret Zoph and Quoc~V Le.
\newblock Neural architecture search with reinforcement learning.
\newblock {\em arXiv preprint arXiv:1611.01578}, 2016.

\end{thebibliography}
}

\renewcommand{\appendixname}{Appendix~\Alph{section}}
\clearpage
\emptythanks
\setcounter{footnote}{0}

\appendix

\section{Appendix}

\subsection{Extensive Experiments on NAS-Bench-201}%

\begin{figure}[h]
    \centering
    \includegraphics[width=\linewidth]{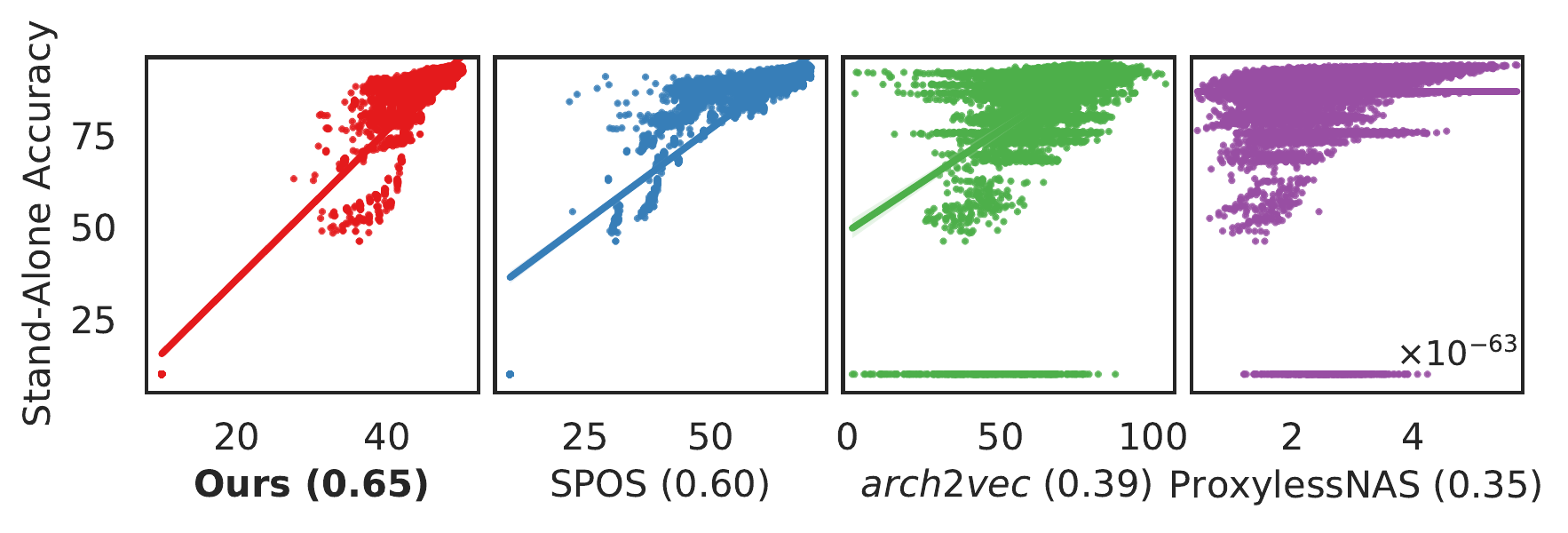}
    \caption{Ranking correlations on all 15625 architectures on NAS-Bench-201~\cite{dong2020nasbench201} on CIFAR-10 compared to SPOS~\cite{guo2020single}, \textit{arch2vec}~\cite{yan2020arch} and ProxylessNAS~\cite{cai2018proxylessnas}.}
    \label{fig:nasbench_correlation_all}
\end{figure}

\begin{figure}[h]
    \centering
    \includegraphics[width=\linewidth]{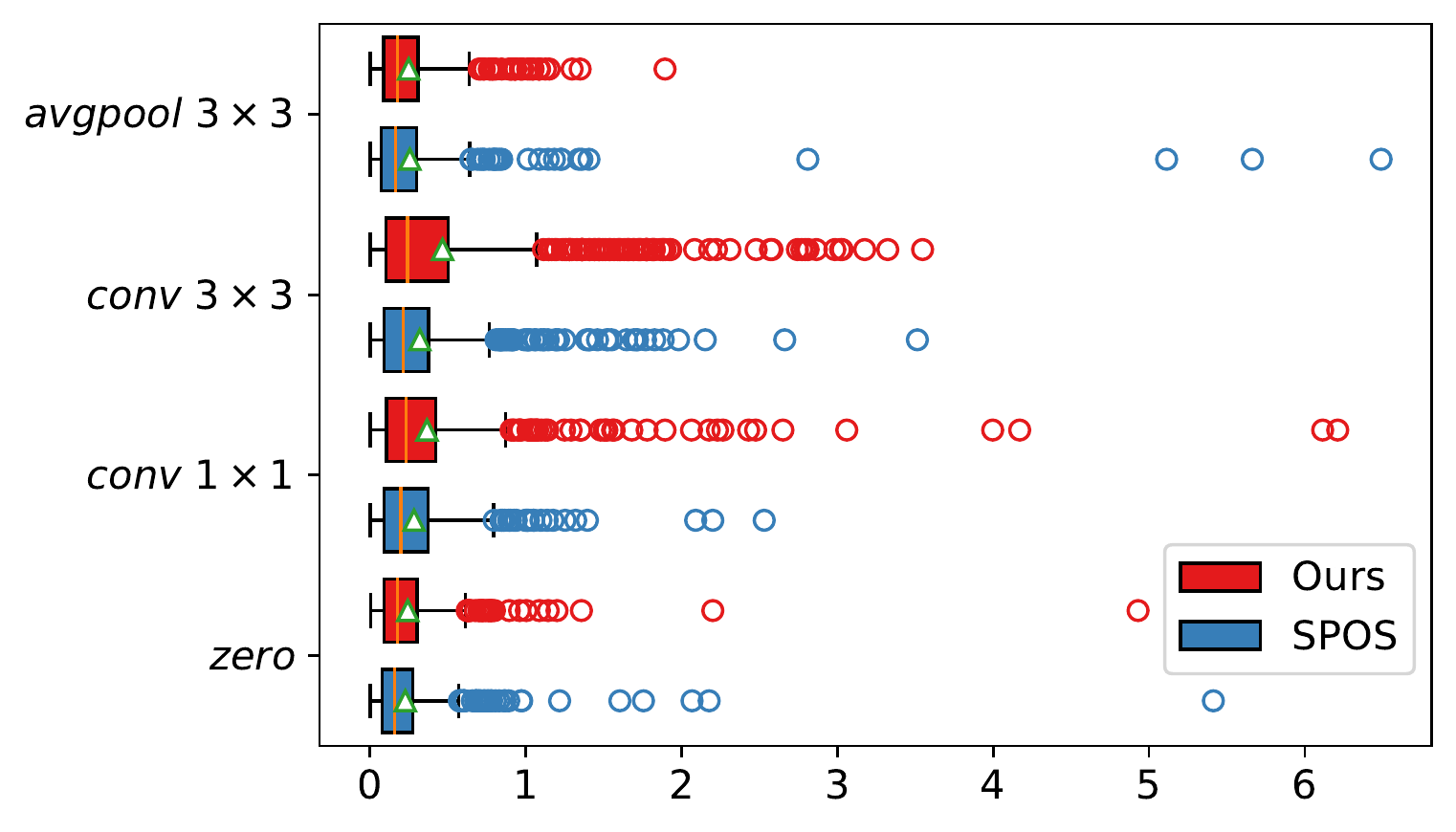}
    \caption{The actual accuracy change (absolute value) of architecture satisfying that the estimated accuracy change is smaller than 0.01 after replacing one operation (\textit{zero, $1\times1$ convolution, $3\times3$ convolution} or \textit{$3\times3$ average pooling}) with \textit{skip connection} for ${\Pi}$-NAS and SPOS~\cite{guo2020single}, respectively.}
    \label{fig:nasbench201_skip_connect}
\end{figure}

\begin{figure}[h]
    \centering
    \includegraphics[width=\linewidth]{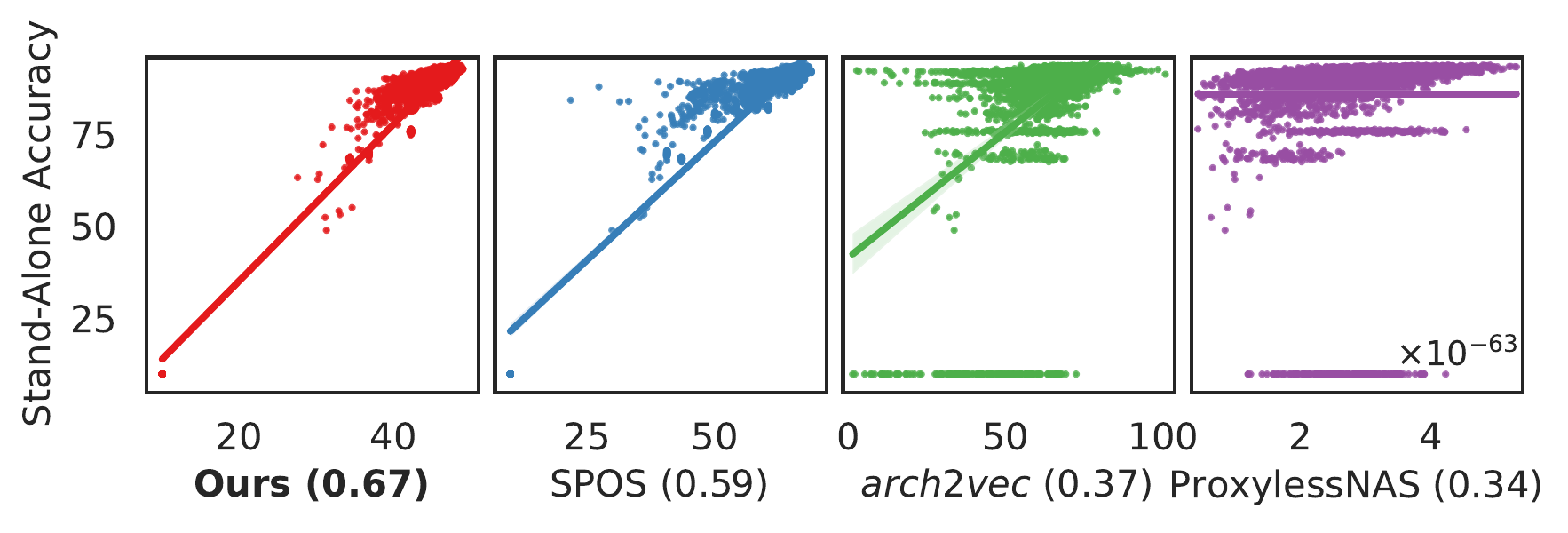}
    \caption{Ranking correlations on 4096 architectures on NAS-Bench-201~\cite{dong2020nasbench201} on CIFAR-10 without \textit{skip connection} operation compared to SPOS~\cite{guo2020single}, \textit{arch2vec}~\cite{yan2020arch} and ProxylessNAS~\cite{cai2018proxylessnas}.}
    \label{fig:nasbench_correlation_wo_skip_connect}
\end{figure}

We also validate our ${\Pi}$-NAS's ranking correlation on all 15625 architectures on NAS-Bench-201~\cite{dong2020nasbench201}. As shown in Figure \ref{fig:nasbench_correlation_all}, our method still outperforms SPOS~\cite{guo2020single}, \textit{arch2vec}~\cite{yan2020arch} and ProxylessNAS~\cite{cai2018proxylessnas}. However, our ${\Pi}$-NAS's advantage over SPOS~\cite{guo2020single} (0.65 \vs 0.60) shrinks compared to the submission's Figure \textcolor{red}{3} (NOT Figure \ref{fig:nasbench_correlation_wo_skip_connect} in this appendix) (0.70 \vs 0.57).

We assume the ranking correlation degradation might be attributed to \textit{skip connection} operation in the search space. To justify this assumption, we provide some visualization in Figure \ref{fig:nasbench201_skip_connect}. Specifically, we replace a non-skip-connection operation (\ie, \textit{zero, $1\times1$ convolution, $3\times3$ convolution} or \textit{$3\times3$ average pooling}) with a \textit{skip connection} for all architectures in the search space. Then, we compare the estimated accuracy change \vs actual accuracy change before and after such a replacement. When an architecture's estimated accuracy change is smaller than 0.01, we plot its actual accuracy change (absolute value) in Figure \ref{fig:nasbench201_skip_connect} for ${\Pi}$-NAS and SPOS, respectively. As shown, there is a significant gap between the estimated accuracy change and actual accuracy change before and after a \textit{skip connection} replacement (\ie, \textbf{before:} $<$ 0.01; \textbf{after:} usually $>$ 1). This visualization indicates \textit{skip connection} operation does hurt ranking correlation for both ${\Pi}$-NAS and SPOS, verifying our assumption.

Previous works have also observed this ranking correlation degradation. As pointed out in \cite{chu2019scarletnas,li2020block}, skip connection can increase supernet scalability in the depth dimension, but it can lead to convergence difficulty of the supernet and unfair comparison of subnets. In addition, our cross-path learning is more prone to this problem since we directly reduce the feature consistency shift between different paths whether with \textit{skip connection} operation or not without special treatment, which can cause the overestimated performances of the architectures containing \textit{skip connection}.

As Figure~\ref{fig:nasbench_correlation_wo_skip_connect} shows, after leaving out the architectures with \textit{skip connection} operation, ${\Pi}$-NAS's advantage recovers. For future work, we will try to develop the scalability of $\Pi$-NAS to solve such limitation.

\vspace{4pt}
\noindent\textcolor{red}{\textbf{There is one more page showing our searched architectures. Don't hesitate to scroll your mouse.}}

\clearpage
\subsection{Model Architectures}%

\begin{figure*}[hb]
    \begin{center}
    \includegraphics[width=0.95\linewidth]{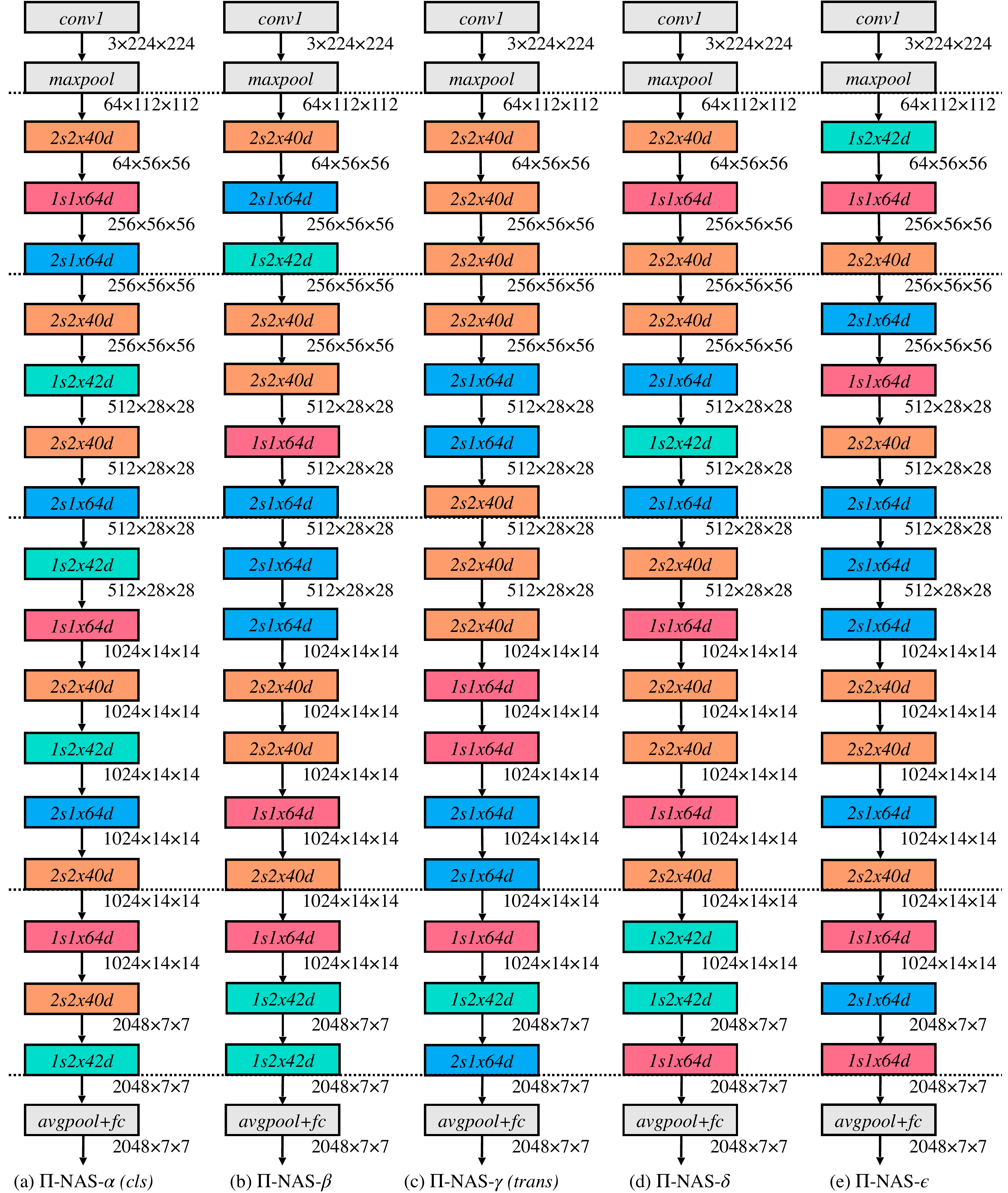}
    \caption{Architectures of $\Pi$-NAS-$\alpha$ (\ie, $\Pi$-NAS-\textit{cls}), $\Pi$-NAS-$\beta$, $\Pi$-NAS-$\gamma$ (\ie, $\Pi$-NAS-\textit{trans}), $\Pi$-NAS-$\delta$ and $\Pi$-NAS-$\epsilon$. They are combinations of \textit{Split-Attention} block with radix $s$, cardinality $x$ and width $d$.}
    \label{fig:arch_detail}
    \end{center}
\end{figure*}

\end{document}